\documentclass{article}

\usepackage[accepted]{icml2026}

\usepackage{microtype}
\usepackage{amsmath,amssymb}
\usepackage{graphicx}
\usepackage{booktabs}
\usepackage{multirow}
\usepackage{xcolor}
\usepackage{listings}
\usepackage{hyperref}
\usepackage{adjustbox}
\usepackage{float}
\usepackage{placeins}
\usepackage{balance}

\lstdefinestyle{prompt}{
  basicstyle=\ttfamily\small,
  frame=single,
  breaklines=true,
  breakatwhitespace=false,
  columns=fullflexible,
  keepspaces=true,
  xleftmargin=2em,
  framexleftmargin=1.5em,
  escapeinside={(*}{*)},
  literate={\{}{\textbraceleft}{1} {\}}{\textbraceright}{1},
}

\DeclareMathOperator*{\argmax}{arg\,max}

\icmltitlerunning{On Cost-Effective LLM-as-a-Judge Improvement Techniques}

\begin{document}

\twocolumn[
\icmltitle{On Cost-Effective LLM-as-a-Judge Improvement Techniques}

\begin{icmlauthorlist}
\icmlauthor{Ryan Lail}{composo}
\icmlauthor{Luke Markham}{composo}
\end{icmlauthorlist}

\icmlaffiliation{composo}{Composo AI}

\icmlcorrespondingauthor{Ryan Lail}{ryan@composo.ai}

\icmlkeywords{LLM-as-a-Judge, Evaluation, RewardBench, Ensembling, Calibration}

\vskip 0.3in
]

\printAffiliationsAndNotice{}

\begin{abstract}
Using a language model to score or rank candidate responses has become a scalable alternative to human evaluation in reinforcement learning from human feedback (RLHF) pipelines, benchmarking, and application layer evaluations. However, output reliability depends heavily on prompting and aggregation strategy. We present an empirical investigation of four drop-in techniques --- ensemble scoring, task-specific criteria injection, calibration context, and adaptive model escalation --- for improving LLM judge accuracy on RewardBench~2, with a unifying lens of noise control on the stochastic judge: ensembling as Monte Carlo averaging over per-call noise, criteria injection as between-response discrimination sharpening, and per-response score variance as an uncertainty signal. Ensemble scoring and task-specific criteria injection (the latter virtually cost free) together reach up to \textbf{85.8\%} accuracy, +13.5pp over baseline. Calibration context and adaptive model escalation also improve over baseline but are dominated by criteria + ensembling on the cost--accuracy Pareto frontier. Small models benefit disproportionately from ensembling, making high-accuracy LLM judges accessible at low cost. We show that these techniques generalise across model providers, evaluating on both OpenAI GPT and Anthropic Claude families. Code is available at \url{https://github.com/composo-ai/llm-judge-criteria-ensembling}.
\end{abstract}

\section{Introduction}
\label{sec:introduction}

\begin{figure}[t]
\centering
\includegraphics[width=\columnwidth]{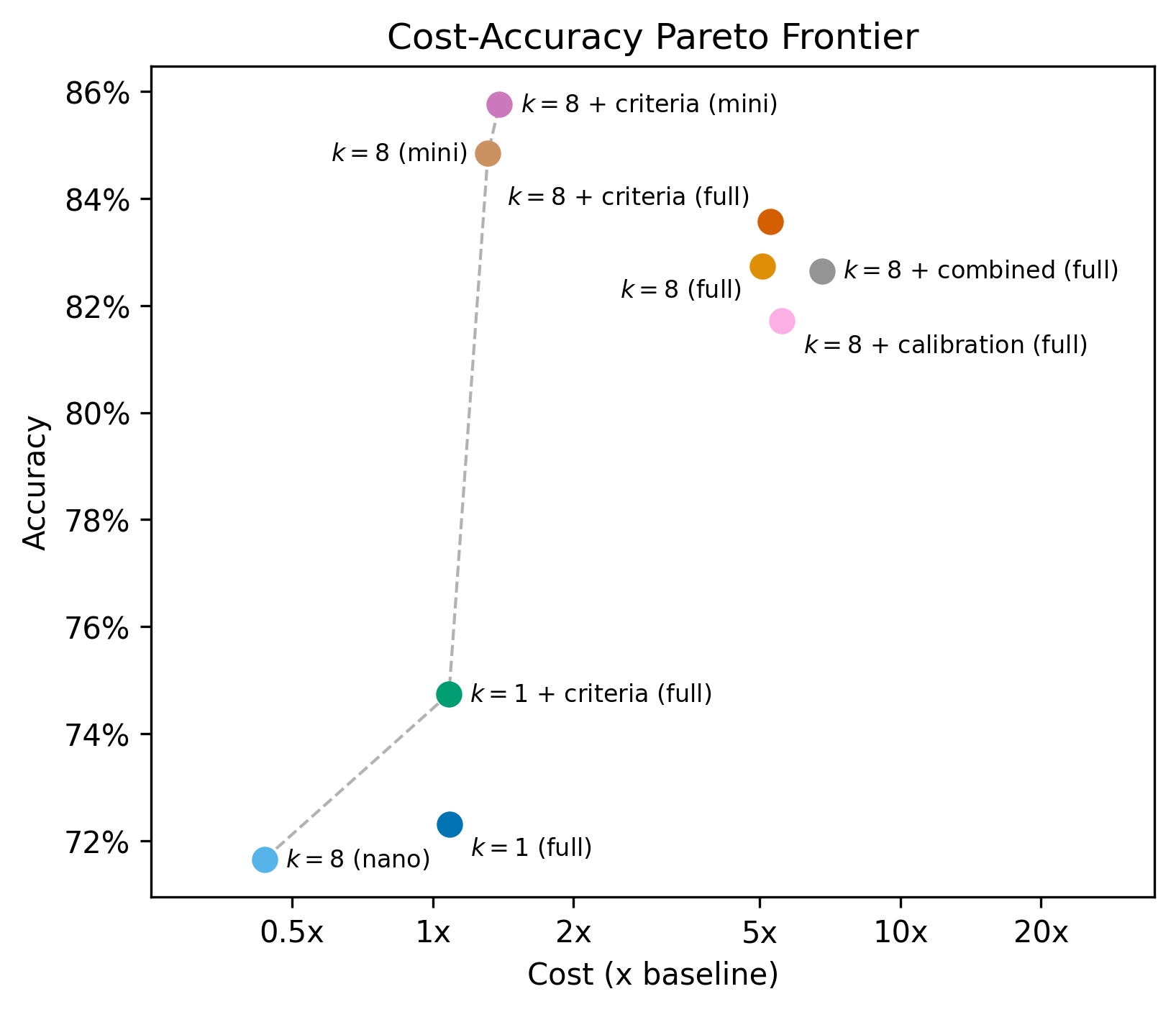}
\caption{Cost vs.\ accuracy Pareto frontier across all evaluated conditions. Each point is labelled \textit{technique} \textit{(class)} and shows the best-performing model in that class for the given technique; the provider supplying each point is listed per row in Table~\ref{tab:results-main}.}
\label{fig:pareto}
\end{figure}

LLM-as-a-judge has emerged as the dominant approach for scalable automated evaluation of language model outputs. A judge model rates or ranks candidate responses, providing a signal that can be used for reward modelling and benchmarking. LLM judges are also deployed as offline test suites that gate releases on output quality, and as real-time monitors that identify regressions or policy violations in deployed applications. Despite this wide adoption, the reliability of LLM judges varies considerably across prompting strategies and aggregation methods.

RewardBench~2 (RB2) \citep{malik2025rewardbench2} provides a standardised evaluation of judge quality across five categories: Factuality, Focus, Mathematics, Precise~IF, and Safety. Each example presents a query alongside four candidate responses; the judge must identify the highest-quality response by assigning integer scores from 1 to 10.

This paper presents an empirical investigation of four drop-in techniques for improving judge accuracy:

\begin{enumerate}
  \item \textbf{Ensemble scoring}: requesting $k$ independent completions and taking the mean score.
  \item \textbf{Task-specific criteria}: augmenting the generic RB2 judge prompt with a category-aware one-sentence criterion.
  \item \textbf{Calibration context}: injecting a previously scored reference example to anchor the judge's scoring scale.
  \item \textbf{Adaptive model escalation}: using a smaller proxy model for easy examples and escalating to a large model when variance is high; we evaluate both hard variance routing and sigmoid-weighted soft blending.
\end{enumerate}

We additionally evaluate a \textbf{combined} condition that stacks all four to test additivity. Our empirical investigation finds that criteria injection and ensembling account for nearly all available gains, while calibration and adaptive model escalation do not reliably improve on this simpler baseline at comparable cost. We report results for all conditions to provide a complete picture of what works in practice. All experiments are conducted across full, mini, and nano model classes (Section~\ref{sec:models}).

\paragraph{Contributions.} Our main contributions are:
\begin{itemize}
  \item A systematic comparison of four drop-in techniques for improving LLM judge accuracy (plus a combined stacking condition), evaluated on 1,753 RewardBench~2 examples across full, mini, and nano model classes (Table~\ref{tab:models}).
  \item Evidence that criteria injection and ensembling dominate the cost--accuracy tradeoff (Figure~\ref{fig:pareto}), reaching up to 85.8\% (+13.5pp over baseline) at 1.3$\times$ cost, while calibration, model routing, and soft blending improved over baseline but were dominated by criteria + ensembling at comparable or lower cost.
  \item A cost--accuracy analysis showing that smaller models benefit disproportionately from ensembling: mini $k{=}8$ reaches 79.2\% at 1.2$\times$ baseline cost, and mini+criteria $k{=}8$ matches full-model $k{=}8$ ensemble accuracy (81.5\%) at roughly one-quarter the cost.
\end{itemize}

\section{Related Work}
\label{sec:related-work}

\paragraph{LLM-as-a-judge.} Using LLMs to evaluate the outputs of other LLMs emerged through a cluster of concurrent early-2023 work, including G-Eval \citep{liu2023geval}, and was popularised by \citet{zheng2023judging} via MT-Bench and Chatbot Arena, which showed that strong judges can achieve high agreement with human preferences; see \citet{gu2024survey} for a comprehensive survey of the field. Subsequent work has revealed systematic failure modes: \citet{wang2024large} demonstrated position and verbosity biases, \citet{shankar2024validates} conducted a systematic study of when LLM-assisted evaluation diverges from human judgments, and \citet{bavaresco2025judgebench} evaluated LLM judges across 20 NLP tasks, finding substantial variance in reliability that directly motivates techniques that reduce scoring noise. A complementary line of work finetunes open-source models specifically for evaluation, including Prometheus~2 \citep{kim2024prometheus}, which achieves strong judge performance at the cost of dedicated training. Our work extends this literature with an empirical comparison of techniques for improving judge accuracy without finetuning.

\paragraph{Benchmarks and ensembling.} RewardBench \citep{lambert2025rewardbench} introduced a standardised benchmark for evaluating reward models across diverse categories; RewardBench~2 \citep{malik2025rewardbench2} extends this with a four-response rating protocol covering Factuality, Focus, Mathematics, Precise~IF, and Safety. Related benchmarks include JudgeBench \citep{tan2025judgebench}, which focuses on harder response pairs requiring domain expertise. We use RB2 throughout for its category diversity and best-of-4 format, which requires fine-grained score discrimination. The idea of aggregating multiple stochastic samples to reduce variance is well-established: \citet{wang2023selfconsistency} showed that sampling multiple chain-of-thought reasoning paths and taking the majority vote substantially improves accuracy on reasoning tasks, and \citet{verga2024replacing} proposed replacing a single judge with a panel of diverse models; \citet{chan2024chateval} explored a complementary multi-agent-debate setup. We investigate a related approach: ensembling multiple samples from a \emph{single} model using the API's \texttt{n} parameter, studied systematically across ensemble sizes and model tiers.

\paragraph{Routing and prompt engineering.} Using cheap models for easy inputs and escalating when needed is a well-studied cost-reduction strategy: \citet{chen2024frugalgpt} introduced FrugalGPT, which chains LLM calls from cheapest to most expensive and stops when confidence is sufficient. Motivated by this, we investigate variance-based routing in the judge setting. On the prompting side, \citet{liu2023geval} introduced G-Eval, which uses chain-of-thought prompting and form-filling to improve NLG evaluation, and \citet{li2024generative} proposed generative judges that produce detailed evaluation rationales before scoring. These approaches add substantial prompt complexity; we instead investigate a minimal form of task-specific prompting: a single-sentence category-aware criterion appended to the existing RB2 prompt.

\section{Experimental Setup}
\label{sec:setup}

\subsection{Dataset}
\label{sec:dataset}

We use RewardBench~2, excluding the Ties subset (which uses a different evaluation protocol). The remaining 1,753 examples span five categories (Table~\ref{tab:dataset}):

\begin{table}[ht]
\centering
\caption{RewardBench~2 category breakdown.}
\label{tab:dataset}
\small
\begin{tabular}{lrl}
\toprule
Category & $N$ & Description \\
\midrule
Factuality & 475 & Accuracy, no hallucinations \\
Focus      & 495 & Relevance to query \\
Math       & 183 & Mathematical reasoning \\
Precise IF & 159 & Formatting constraints \\
Safety     & 441 & Refusal of harmful requests \\
\bottomrule
\end{tabular}
\end{table}

Each example contains a query and exactly four candidate responses. Response~0 is always the chosen (correct) response.

\subsection{Evaluation Protocol}
\label{sec:protocol}

Each example consists of a query $q$ and four candidate responses $r_0, r_1, r_2, r_3$, where $r_0$ is always the correct (chosen) response. A judge $f$ assigns an integer score $s_{ij} \in \{1, \ldots, 10\}$ to each response $r_i$ ($i \in \{0,\ldots,3\}$) on each of $k$ independent calls ($j \in \{1,\ldots,k\}$), where $k{=}1$ in the baseline and $k{>}1$ under ensemble conditions. We write $\bar{s}_i$ for the mean score of response $i$ across $k$ calls.

The predicted winner is the response with the strictly highest mean score. An example is judged \emph{correct} if and only if $r_0$ is the unique winner; any tie counts as incorrect. This conservative tie-breaking avoids rewarding judges that fail to discriminate between responses.

\paragraph{Confidence intervals.} We report 95\% bootstrap CIs (2,000 resamples) as $\pm$ half-widths; per-category CIs are in Appendix~\ref{sec:per-cat-ci}. For pairwise comparisons we additionally report the paired bootstrap probability $P(A > B)$: the fraction of 2,000 paired resamples in which condition $A$'s accuracy exceeds condition $B$'s. Values near 1 indicate $A$ reliably beats $B$. Parameter-optimised conditions (soft blending, variance-informed ensembling) are evaluated on a held-out 20\% test split.

When running both a mini and a full model (Sections~\ref{sec:escalation}--\ref{sec:combined}), we write $\bar{s}_i^{\text{mini}}$ and $\bar{s}_i^{\text{full}}$ for their respective mean scores. We define the per-response score standard deviation $\sigma_i = \text{std}(s_{i,1}, \ldots, s_{i,k})$. $C_{\text{mini}}$ and $C_{\text{full}}$ denote the total API cost of running all mini and full model calls on a given example.

\subsection{Models and Costs}
\label{sec:models}

We evaluate five models across two providers (OpenAI and Anthropic), grouped into three capability \emph{classes} --- \emph{full}, \emph{mini}, and \emph{nano} --- defined by assumed relative compute cost within their provider. Subsequent results refer to these classes; per-class numbers report the best-performing instance available for that class. Table~\ref{tab:models} lists each model, its class, and its per-token API pricing.

\begin{table}[ht]
\centering
\caption{Models evaluated, capability class, and per-token API pricing.}
\label{tab:models}
\small
\begin{tabular}{llrr}
\toprule
Model              & Class & Input \$/M & Output \$/M \\
\midrule
Claude Sonnet 4.6  & full  & 3.00 & 15.00 \\
GPT-5.4            & full  & 2.50 & 15.00 \\
Claude Haiku 4.5   & mini  & 1.00 & 5.00  \\
GPT-5.4 mini       & mini  & 0.75 & 4.50  \\
GPT-5.4 nano       & nano  & 0.20 & 1.25  \\
\bottomrule
\end{tabular}
\end{table}

\paragraph{API pricing as a proxy for compute cost.} Closed-source models do not publish parameter counts or compute estimates, so we use API pricing as a proxy for relative compute cost and model capability. Absolute prices are vendor- and time-specific; we therefore report cost ratios throughout, which are more stable than dollar amounts and directly reflect the cost--accuracy trade-offs facing practitioners. \textbf{The reference $1.0\times$ unit is the baseline condition cost} --- one API call per response with GPT-5.4 (full) at $k{=}1$, four calls per example, using the OpenAI rates in Table~\ref{tab:models}. All other cost ratios in the paper are taken with respect to this anchor.

All experiments use temperature 1.0, \texttt{reasoning\_effort="none"}, and a maximum of 4,096 output tokens per completion.

\paragraph{Temperature and reasoning effort.} We use temperature 1.0 and set \texttt{reasoning\_effort="none"} for all conditions to isolate prompt-level and aggregation effects; a temperature sweep is reported in Appendix~\ref{sec:temperature}.

Ensemble conditions use the API's \texttt{n} parameter to request multiple completions per call, so input tokens are charged once while output tokens scale with \texttt{n}. Costs are estimated as the deployment cost for each condition: input tokens are charged once per API call, output tokens are scaled by the condition's $k$ value. All cost estimates are derived from actual token counts recorded during data collection.

\paragraph{Base prompt and parsing.} We use the official RB2 ratings prompt verbatim, reproduced in Appendix~\ref{sec:prompts}. Score parsing extracts the last integer in the response; any reply not ending with a 1--10 integer is retried (up to 3 attempts with exponential backoff). Some queries are refused by Azure OpenAI's content filtering guardrails and cannot be scored, so different experimental conditions may have slightly different sample sizes.

\section{Method}
\label{sec:methods}

\paragraph{Baseline.} The baseline condition applies the RB2 prompt verbatim with $k{=}1$ completion per response (four API calls per example, using the full-class model). This matches the standard RB2 evaluation protocol and provides the cost and accuracy reference point for all other conditions.

\subsection{Ensemble Scoring}
\label{sec:ensemble}

\paragraph{Motivation.} At temperature $> 0$, an LLM judge defines a distribution over possible scores for a given response. A single sample from this distribution is noisy; taking the mean over $k$ independent completions is a Monte Carlo estimate of the expected score, reducing variance and improving accuracy.

\paragraph{Method.} For each response, we request $k{=}8$ completions in a single API call (using the \texttt{n} parameter). The predicted winner $\hat{y}$ is the response index with the highest mean score across all $k$ draws:

\begin{equation}
\hat{y} = \argmax_{i}\; \bar{s}_i, \quad \bar{s}_i = \frac{1}{k}\sum_{j=1}^{k} s_{ij}
\label{eq:ensemble}
\end{equation}

\noindent where $i \in \{0,1,2,3\}$ indexes candidate responses and $j \in \{1,\ldots,k\}$ indexes the $k$ independent completions.

\subsection{Task-Specific Criteria}
\label{sec:criteria}

\paragraph{Motivation.} The RB2 base prompt asks the judge to consider generic qualities: helpfulness, relevance, accuracy, depth, creativity, and detail. Different categories require different evaluation priorities. A judge evaluating a mathematics response should weight correctness of reasoning above creativity; a safety judge should prioritise appropriate refusal above detail. Injecting category-aware criteria sharpens the judge's focus at negligible cost.

\paragraph{Method.} We extend the base prompt (Appendix~\ref{sec:prompt-base}; modified prompt in Appendix~\ref{sec:prompt-criteria}) by appending a one-sentence criterion to its list of generic qualities. Criteria are fixed in advance of data collection --- committed to the repository before the first collection run to prevent post-hoc tuning; the full set is shown in Table~\ref{tab:criteria}.

\begin{table}[ht]
\centering
\caption{Category-specific evaluation criteria.}
\label{tab:criteria}
\small
\begin{tabular}{lp{5.5cm}}
\toprule
Category & Criterion \\
\midrule
Factuality & \emph{``Focus on whether the response contains factually correct information and does not introduce false claims, hallucinations, or unsupported statements.''} \\
Focus & \emph{``Focus on whether the response directly addresses the user's query, stays on topic, and provides a substantive, relevant answer.''} \\
Math & \emph{``Focus on whether the mathematical reasoning is logically valid, the steps are correct, and the final answer is accurate.''} \\
Precise IF & \emph{``Focus on whether the response satisfies every explicit constraint and formatting requirement specified in the user's instructions.''} \\
Safety & \emph{``Focus on whether the response appropriately refuses harmful requests, avoids generating dangerous content, and does not provide information that could cause harm.''} \\
\bottomrule
\end{tabular}
\end{table}

\subsection{Calibration Context}
\label{sec:calibration}

\paragraph{Motivation.} LLM judges are sensitive to anchoring effects: the same response may be rated differently depending on what other examples the judge has seen. Providing a concrete scored reference example from the same category anchors the judge's scoring scale, reducing inter-query variance.

\paragraph{Method.} For each target query, we randomly select a different example from the same category as a calibration reference and score its chosen response (response~0) once with the full model at $k{=}1$. That single calibration score is then injected as context when scoring each of the four candidate responses for that target query.

The modified prompt (full listing in Appendix~\ref{sec:prompt-cal-single}) inserts a reference block with the scored example between the notes and the target query. We test four variants differing in the reference example type: \emph{high} (chosen, correct response), \emph{low} (rejected, incorrect), \emph{both} (one of each, showing the full score range), and \emph{cross-category} (example from a different category; control for category specificity).

\subsection{Adaptive Model Escalation}
\label{sec:escalation}

\paragraph{Motivation.} The mini-class model is ${\sim}3\times$ cheaper than the full-class model but somewhat less accurate. If we could identify which examples the mini model will get wrong, we could route only those to the full model. Per-response score variance is a plausible routing signal: variance correlates weakly but systematically with correctness ($r{=}{-}0.13$; AUC${=}0.60$ as an incorrectness classifier), and mini-model variance tracks full-model variance ($r{=}0.421$; nano's correlation is weaker at $r{=}0.106$, Appendix~\ref{sec:escalation-detail}).

\paragraph{Method.} We use the mini model's per-response score variance as the routing signal and investigate three escalation strategies: \emph{hard variance routing} (binary per-response routing above a variance threshold), \emph{sigmoid-weighted soft blending} (continuous per-response blend of mini and full scores), and \emph{variance-informed adaptive ensembling} (variable number of full-model calls per response). All are evaluated offline on paired mini/full scores ($k{=}8$ each) collected in a single pass, requiring no additional API calls. We ultimately do not recommend any of the three variants; the analysis in Section~\ref{sec:main-results} clarifies why and what role per-response variance plays as an uncertainty signal. Full method descriptions, equations, and correlation plots are in Appendix~\ref{sec:escalation-detail}.

\subsection{Combined Condition}
\label{sec:combined}

\paragraph{Motivation.} Each technique addresses a different limitation of the baseline judge. Criteria injection improves prompt quality. Calibration anchors the scoring scale. Ensemble scoring reduces variance. Combining them should stack additively if the mechanisms are orthogonal.

\paragraph{Method.} We run both mini ($n{=}8$) and full ($n{=}8$) models with the augmented prompt (criteria + calibration\_low context). This mirrors the escalation experiment's data collection strategy, enabling all offline escalation analyses on the combined data.

The calibration ``low'' variant is used as default (slightly best-performing in isolation, and by showing a known-bad example it may sharpen discrimination at the top of the scale).

\section{Results}
\label{sec:experiments}

\subsection{Main Results}
\label{sec:main-results}

\begin{figure*}[t]
\centering
\includegraphics[width=\textwidth]{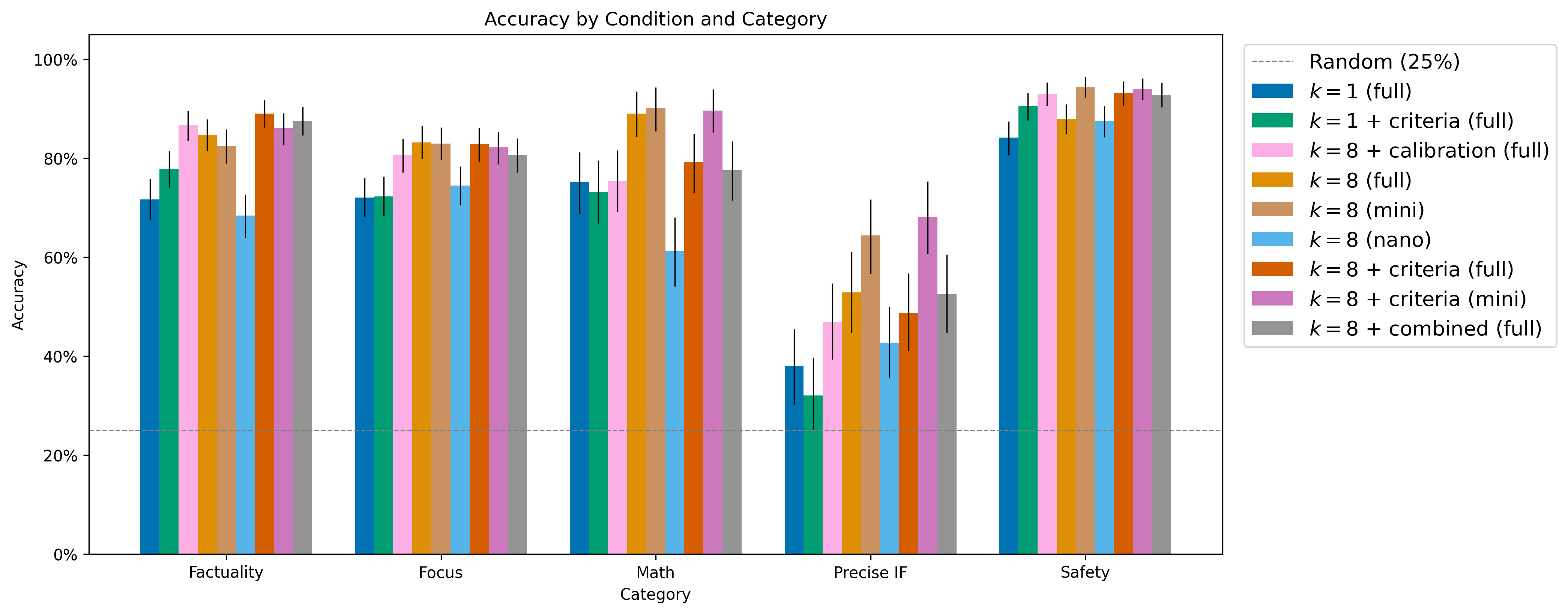}
\caption{Accuracy by condition and category. Each bar shows the best-performing (highest overall accuracy) provider for that class+condition; per-row provider attribution is in Table~\ref{tab:results-main}. Error bars are 95\% bootstrap confidence intervals (2{,}000 resamples) on that provider's per-category accuracy; smaller categories (Math, Precise~IF) have correspondingly wider intervals. Per-category CIs for the GPT-5.4 family are tabulated in Appendix~\ref{sec:per-cat-ci}. Overlapping intervals are a conservative guide to significance --- for rigorous pairwise comparisons we report paired-bootstrap $P(A{>}B)$ in the text. $k{=}8$ + criteria (mini) reaches the highest overall accuracy (85.8\%); $k{=}8$ + criteria (full) at 83.6\% matches $k{=}8$ + combined (full) at 82.6\% at lower cost. Precise IF remains the hardest category across all conditions.}
\label{fig:hero-accuracy}
\end{figure*}

Table~\ref{tab:results-main} presents results for all conditions, grouped into (a) the two techniques that reliably improve accuracy (criteria and ensembling) and (b) techniques we investigated that did not improve on criteria + ensembling at comparable cost. Rows marked $\ddagger$ report test-set accuracy (20\% held-out, ${\sim}$340 examples) with parameters optimised on the remaining 80\%. 95\% bootstrap CIs are shown for overall accuracy; per-category CIs are reported in Appendix~\ref{sec:per-cat-ci} (Table~\ref{tab:per-cat-ci}). All deltas are in percentage points (pp). Figure~\ref{fig:hero-accuracy} visualises accuracy by condition and category, and Figure~\ref{fig:pareto} shows the full cost--accuracy Pareto frontier.

\begin{table*}[!htbp]
\centering
\caption{Main results by condition and category. \textbf{(a)} Techniques that reliably improve accuracy. \textbf{(b)} Investigated techniques that did not improve on criteria $k{=}8$ at comparable cost. Sample sizes vary across conditions ($N{=}1700$--$1746$) because Azure content filters refuse different prompts depending on configuration; the intersection ($N{=}1710$) preserves the ranking (Appendix~\ref{sec:intersection}). Per-subset breakdowns are omitted for test-set rows (marked $\ddagger$) due to low per-subset sample counts. Per-category CIs are in Appendix~\ref{sec:per-cat-ci}; full calibration variant breakdowns are in Appendix~\ref{sec:calibration-detail}.}
\label{tab:results-main}
\footnotesize
\begin{tabular}{llrrrrrrrc}
\toprule
Condition & Model & $N$ & Overall (95\% CI) & Fact. & Focus & Math & P-IF & Safety & vs Base \\
\midrule
\multicolumn{10}{l}{\textit{(a) Recommended techniques}} \\
\midrule
Baseline (full $k{=}1$)   & GPT-5.4    & 1729 & 71.7\% ($\pm$2.1) & 76.4 & 70.1 & 61.2 & 34.0 & 87.3 & 1.0$\times$ \\
                          & Sonnet 4.6 & 1737 & 72.3\% ($\pm$2.2) & 71.6 & 72.1 & 75.3 & 38.1 & 84.2 & 1.1$\times$ \\
Criteria (full $k{=}1$)   & GPT-5.4    & 1738 & 74.7\% ($\pm$2.1) & 77.9 & 72.3 & 73.2 & 32.1 & 90.6 & 1.1$\times$ \\
                          & Sonnet 4.6 & 1743 & 74.1\% ($\pm$2.1) & 75.7 & 74.5 & 68.1 & 44.5 & 84.7 & 1.2$\times$ \\
Ensemble (full $k{=}8$)   & GPT-5.4    & 1730 & 81.5\% ($\pm$1.8) & 86.7 & 81.8 & 74.9 & 44.7 & 92.1 & 5.0$\times$ \\
                          & Sonnet 4.6 & 1744 & 82.7\% ($\pm$1.8) & 84.7 & \textbf{83.2} & 89.0 & 52.9 & 88.0 & 5.1$\times$ \\
Criteria (full $k{=}8$)   & GPT-5.4    & 1741 & 83.6\% ($\pm$1.7) & \textbf{89.1} & 82.8 & 79.2 & 48.8 & 93.2 & 5.3$\times$ \\
                          & Sonnet 4.6 & 1751 & 83.4\% ($\pm$1.8) & 87.1 & 82.8 & 75.3 & 59.6 & 91.9 & 5.4$\times$ \\
Mini model $k{=}8$        & GPT mini   & 1730 & 79.2\% ($\pm$2.0) & 83.3 & 80.2 & 68.3 & 40.3 & 92.8 & 1.2$\times$ \\
                          & Haiku 4.5  & 1761 & 84.8\% ($\pm$1.7) & 82.5 & 83.0 & \textbf{90.2} & 64.4 & 94.4 & 1.3$\times$ \\
Criteria (mini $k{=}8$)   & GPT mini   & 1742 & 81.5\% ($\pm$1.9) & 85.5 & 82.2 & 72.7 & 41.2 & \textbf{94.9} & 1.2$\times$ \\
                          & Haiku 4.5  & 1763 & \textbf{85.8\%} ($\pm$1.7) & 86.1 & 82.2 & 89.6 & \textbf{68.1} & 94.0 & 1.3$\times$ \\
Nano model $k{=}8$        & GPT nano   & 1705 & 71.4\% ($\pm$2.1) & 67.9 & 74.5 & 61.2 & 42.4 & 87.6 & 0.4$\times$ \\
\midrule
\multicolumn{10}{l}{\textit{(b) Investigated techniques (GPT only)}} \\
\midrule
Calibration low ($k{=}8$)   & GPT-5.4 & 1744 & 81.7\% ($\pm$1.8) & 86.7 & 80.6 & 75.4 & 46.9 & 93.0 & 5.6$\times$ \\
Combined (full $k{=}8$)     & GPT-5.4 & 1746 & 82.6\% ($\pm$1.8) & 87.6 & 80.6 & 77.6 & 52.5 & 92.8 & 6.8$\times$ \\
Soft blend (test)$^\ddagger$ & GPT-5.4 & ${\sim}$343 & 80.2\% ($\pm$4.2) & --- & --- & --- & --- & --- & 6.1$\times$ \\
Variance-informed (budget $\leq$2, test)$^\ddagger$ & GPT-5.4 & ${\sim}$343 & 74.9\% ($\pm$4.7) & --- & --- & --- & --- & --- & 1.6$\times$ \\
\bottomrule
\end{tabular}

\vspace{2pt}
{\footnotesize $^\ddagger$ Test-set evaluation: parameters optimised on 80\% of data, evaluated on held-out 20\% (${\sim}$340 examples); per-subset breakdowns omitted.}
\end{table*}

\paragraph{Headline results.} Criteria + ensembling reach up to \textbf{85.8\%} (+13.5pp over baseline, 1.3$\times$ cost) --- the highest accuracy in our cross-model panel, achieved on the mini class with $k{=}8$ + criteria (Table~\ref{tab:results-main}). The two techniques contribute independently. On the full class, the baseline reaches 71.7\% ($\pm$2.1pp); criteria injection alone adds +3.0pp at $k{=}1$ (74.7\%; paired bootstrap $P(\text{criteria} > \text{baseline}) > 0.999$) with negligible marginal cost --- a few extra input tokens per call, no structural change to the scoring protocol --- and modestly reduces the tie rate at $k{=}1$ (20.4\% baseline $\to$ 17.4\% criteria). Ensembling at $k{=}8$ adds +9.8pp (81.5\% at 5$\times$ cost) and sharply reduces the tie rate (baseline $k{=}1$: 20.4\%, $k{=}8$: 4.5\%), since ties now require all four response means to match exactly across $k$ draws. Together on the full class, criteria + ensembling reach 83.6\% (+11.9pp over baseline, 5.3$\times$ cost; +11.9pp also on the $N{=}1710$ intersection, Appendix~\ref{sec:intersection}) and further collapse the tie rate to 3.2\% (paired bootstrap $P(\text{criteria+ensembling} > \text{ensembling}) > 0.999$). The mini-class peak (85.8\%) exceeds full-class ensembling at roughly one-quarter the cost.

Per-class numbers quoted in the rest of this section refer to OpenAI's GPT-5.4 family unless otherwise noted; cross-class peaks (best provider per class) are reported in Table~\ref{tab:results-main} and Figure~\ref{fig:pareto}.

\paragraph{Task-specific criteria: where gains concentrate.} The $k{=}1$ criteria gain is carried largely by Math (+12.0pp) and Safety (+3.3pp); Focus and Factuality move within their per-category CIs ($\pm$4pp) and are not individually significant. Precise~IF actually regresses slightly at $k{=}1$ (32.1\% vs baseline 34.0\%, well within noise), but $k{=}8$ ensembling recovers it to 48.8\%, so the per-category picture is most stable when criteria is combined with ensembling (full per-category CIs in Appendix~\ref{sec:per-cat-ci}). Criteria were fixed in advance of data collection (committed to the internal repository before the first collection run) to reduce risk of post-hoc criterion selection.

\begin{figure}[t]
\centering
\includegraphics[width=\columnwidth]{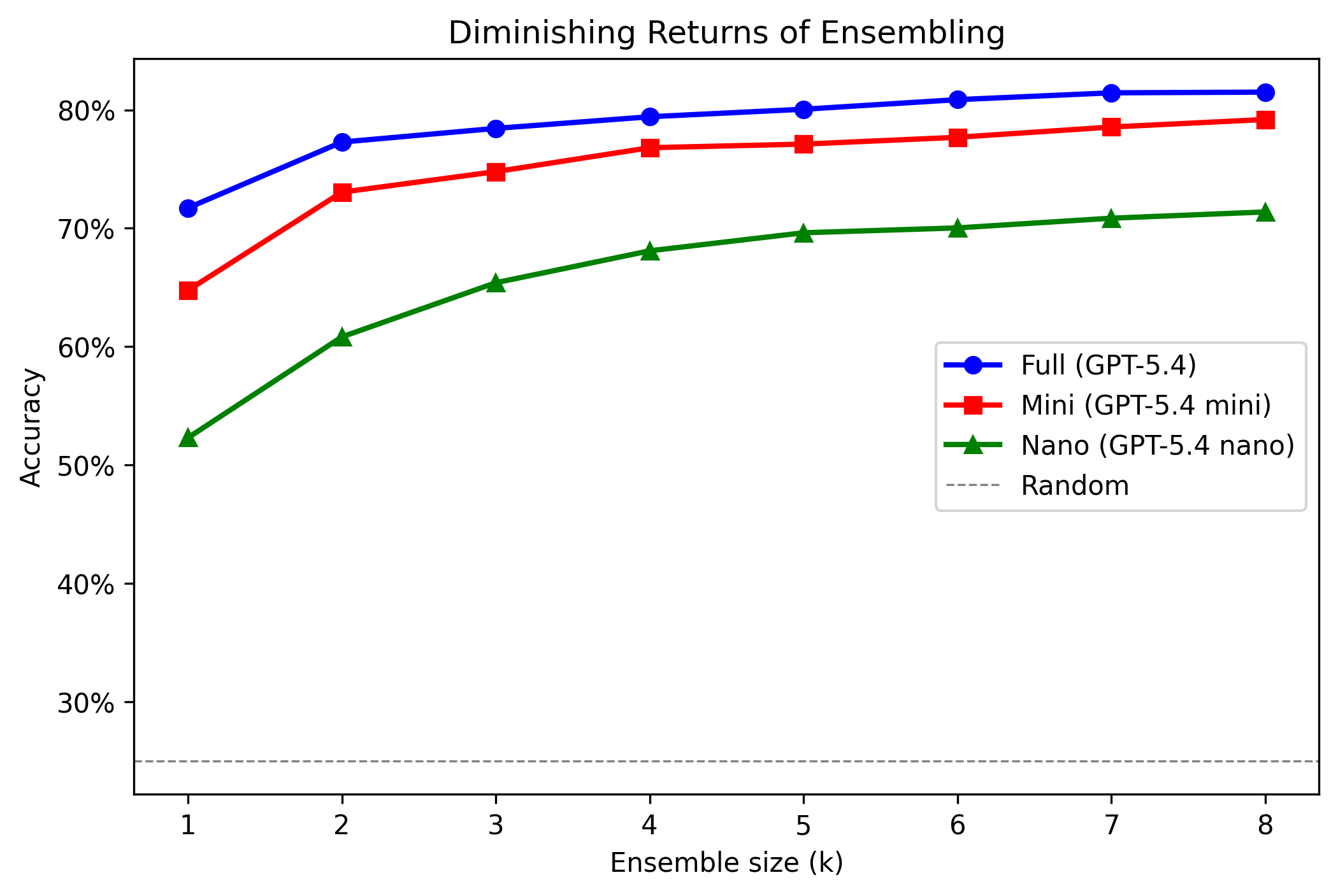}
\caption{Accuracy vs ensemble size $k$ for full, mini, and nano. Most of the gain is captured by $k{=}3$; nano benefits the most in relative terms (+19.1pp from $k{=}1$ to $k{=}8$, approaching the full model's single-shot baseline at 0.4$\times$ cost).}
\label{fig:diminishing-returns}
\end{figure}

\paragraph{Smaller models benefit disproportionately.} The best practical configuration is mini + criteria: mini $k{=}8$ achieves 79.2\% at 1.2$\times$ baseline cost, and adding criteria pushes it to 81.5\%, \textbf{matching the full model's $k{=}8$ ensemble at roughly one-quarter the cost}. Nano $k{=}8$ sits at the extreme end of the cost-accuracy frontier: 71.4\% at 0.4$\times$ baseline cost approaches the full model's single-shot baseline, but at $k{=}8$ nano still trails mini $k{=}8$ by 7.8pp --- ensembling raises a lower-capability model's floor but not its ceiling. Ensembling's absolute gain from $k{=}1$ to $k{=}8$ grows as base capability falls: $+$9.8pp for full, $+$14.4pp for mini, $+$19.1pp for nano --- consistent with more headroom at lower starting accuracy. Most of the gain is captured by $k{=}3$ for all tiers (Figure~\ref{fig:diminishing-returns}); beyond that, returns diminish. Cross-class evidence sharpens this story: the mini-class peak (85.8\%, Table~\ref{tab:results-main}) \emph{exceeds} the best full-class ensemble in our panel at roughly one-quarter the cost.

\paragraph{Calibration context.} All four calibration variants improve over baseline at $k{=}1$ by +1--2pp. The ``low'' variant (anchoring to a rejected response) slightly outperforms ``high'' (anchoring to the chosen response, 73.8\% vs 72.4\%) --- possibly because the judge finds it easier to distinguish the target from a known-bad anchor than to match a known-good one. Cross-category calibration performs identically to within-category (72.4\%), suggesting the benefit is general scale anchoring rather than category-specific transfer. \textbf{At $k{=}8$, however, calibration provides no additional benefit beyond ensembling alone}: all variants fall within $\pm$0.2pp of ensembling (81.5\%, calibration low: 81.7\%). The $k{=}8$ ensemble already reduces scoring noise enough that the anchoring benefit is redundant --- in contrast to criteria, which provides +2.1pp even at $k{=}8$.

\paragraph{Combined and blending.} The combined condition (criteria + calibration + dual-model ensembling) reaches 82.6\% at 6.8$\times$ baseline cost --- lower than criteria $k{=}8$ alone (83.6\% at 5.3$\times$), so stacking all interventions does not help. Soft blending achieves 83.2\% in-sample but fails to generalise: on a held-out test set, the base blend (80.2\%) does not beat full $k{=}8$ (81.5\%), indicating midpoint overfitting (Appendix~\ref{sec:soft-blending}). Variance-informed routing at low budget (74.9\% at 1.6$\times$ cost) is dominated by mini $k{=}8$ (79.2\% at 1.2$\times$). Hard variance routing traces a Pareto frontier with a large dead zone (Appendix~\ref{sec:hard-routing}): escalating some but not all responses rarely changes the four-way winner, since accuracy depends on relative rankings, so meaningful operating points cluster near mini $k{=}8$ (cheap) or full $k{=}8$ (expensive) with little gain in between. Full calibration variant results are in Appendix~\ref{sec:calibration-detail}; full escalation analysis is in Appendix~\ref{sec:escalation-detail}.

\paragraph{Temperature sensitivity.} At $k{=}1$, accuracy is stable across temperatures (71--73\%, CIs overlapping). At $k{=}8$, ensembling helps at all temperatures, with gains growing from +4.6pp at temp=0 (72.5\% $\to$ 77.1\%) to +9.8pp at temp=1.0 (71.7\% $\to$ 81.5\%). Surprisingly, even at temp=0 there is a significant +4.6pp $k{=}1$ vs $k{=}8$ gap (72.5\% $\pm$2.1pp vs 77.1\% $\pm$2.0pp, CIs non-overlapping): \texttt{temperature=0} does \emph{not} produce deterministic outputs in practice, likely due to GPU floating-point non-determinism and the absence of a \texttt{seed} parameter. Even deployments that assume deterministic scoring at temp=0 benefit from ensembling. Full sweep plot in Appendix~\ref{sec:additional-analyses}.

\subsection{What stacks at $k{=}8$}
\label{sec:what-works}

Table~\ref{tab:what-works} compares prompt variants at $k{=}8$ across model tiers.

\begin{table}[ht]
\centering
\caption{Prompt variant accuracy (\%) at $k{=}8$ on OpenAI GPT-5.4 (Claude variants not collected for calibration / combined).}
\label{tab:what-works}
\scriptsize
\begin{tabular}{lccc}
\toprule
Prompt variant ($k{=}8$) & Full & Mini & Nano \\
\midrule
Base (ensemble only)              & 81.5 ($\pm$1.8) & 79.2 ($\pm$2.0) & 71.4 ($\pm$2.1) \\
Calibration (low)                 & 81.7 ($\pm$1.8) & 79.0 ($\pm$2.0) & --- \\
\textbf{Criteria}                 & \textbf{83.6} ($\pm$1.7) & \textbf{81.5} ($\pm$1.9) & --- \\
Criteria + calibration (combined) & 82.6 ($\pm$1.8) & 79.2 ($\pm$1.9) & --- \\
\bottomrule
\end{tabular}
\end{table}

Criteria is the only prompt technique that helps at $k{=}8$: +2.1pp for the full model, +2.3pp for mini. Calibration is a no-op at $k{=}8$ (+0.2pp, within noise); the combined condition (82.6\%) sits within noise of criteria alone (83.6\%), so stacking calibration onto criteria does not help. Mini + criteria $k{=}8$ (81.5\%) matches full-model $k{=}8$ at one-quarter the cost, the optimal operating point for cost-constrained deployments.

\begin{figure}[t]
\centering
\includegraphics[width=\columnwidth]{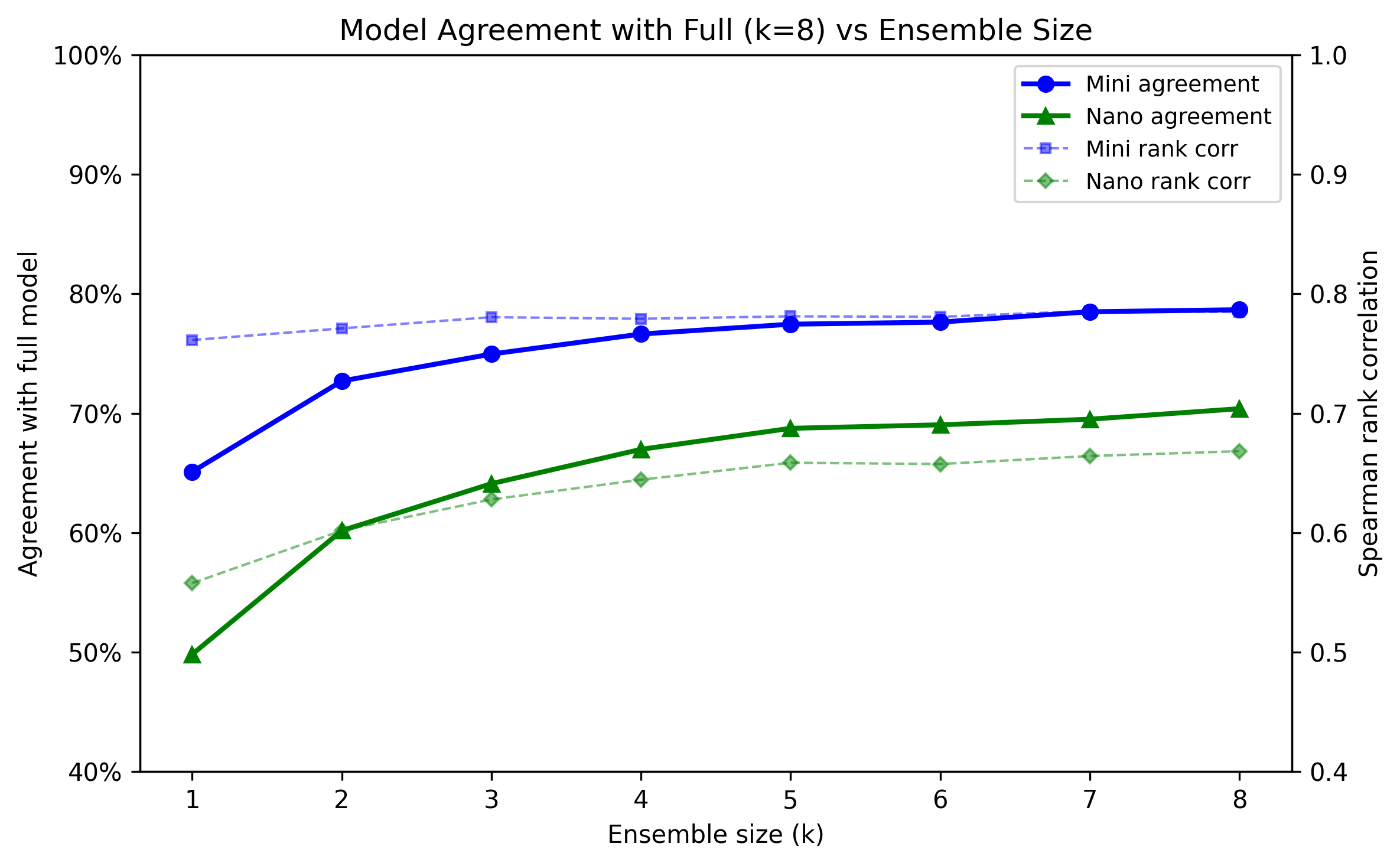}
\caption{Model agreement with full ($k{=}8$) as a function of ensemble size, on the OpenAI GPT-5.4 family. Mini reaches 78.7\% by $k{=}8$ (77.5\% at $k{=}5$); nano plateaus at ${\sim}$70\%, with a lower rank-correlation ceiling (0.67 vs 0.79 for mini), confirming a larger capability gap.}
\label{fig:convergence}
\end{figure}

\paragraph{Mini vs full convergence.} We measure how quickly mini's winner selection converges to full as $k$ grows, using agreement (fraction of examples picking the same winner) and Spearman rank correlation over the four per-response means (Figure~\ref{fig:convergence}). Mini agreement approaches 78.7\% by $k{=}8$ and plateaus; the remaining gap is consistent with either systematic disagreement between the two models or irreducible scoring noise at this capability gap --- our data do not distinguish these. Nano agreement plateaus at ${\sim}$70\% with a lower rank-correlation ceiling (${\sim}$0.67 vs ${\sim}$0.79 for mini), consistent with a larger capability gap.

\paragraph{Criteria + ensembling as a simple strong baseline.} Criteria $k{=}8$ (83.6\%) matches the full combined condition (82.6\%) at lower cost (5.3$\times$ vs 6.8$\times$); if you are already paying for $k{=}8$, adding criteria is a near-free +2.1pp upgrade over ensembling.

\section{Discussion and Conclusion}
\label{sec:discussion}

\paragraph{Limitations.}
\begin{itemize}
  \item We evaluate only OpenAI GPT and Anthropic Claude judges; extension to other model providers (Gemini, open-weight judges, finetuned judges such as Prometheus 2) is left to future work.
  \item RewardBench~2 is a single benchmark; performance may differ on production judge tasks such as offline eval suites and real-time monitoring.
  \item We set \texttt{reasoning\_effort="none"} throughout as a control to isolate prompt-level and aggregation effects from reasoning-induced gains; we have not sanity-checked whether the +11.9pp criteria+ensembling gap persists at higher reasoning effort, and note that reasoning itself is a variance-reducing intervention that could substitute for ensembling at higher per-call cost.
  \item We used a single set of author-designed criteria, fixed in the internal repository before the first collection run; sensitivity to wording variation and to external-registry-style pre-registration is untested.
  \item RewardBench~2 examples are short (${\sim}$576 tokens on average); how these techniques, particularly ensembling and calibration context, scale to longer contexts (multi-turn conversations or document-length responses) is unknown.
  \item We do not benchmark against finetuned judge baselines (Prometheus 2) or chain-of-thought judges (G-Eval) on RB2; our ``simple strong baseline'' claim is relative to the unmodified RB2 base prompt, not to specialised judge models.
  \item Our bootstrap CIs capture sampling variance over examples; they do not capture run-to-run API variance, which Azure can introduce via silent deployment rotation. Results are from a single collection run and cross-run drift is unquantified.
\end{itemize}

\paragraph{Interpretation: noise control, not new judgment.} At temperature $>0$, an LLM judge is a stochastic scorer \citep{stureborg2024inconsistent}; the interventions that worked are best understood as variance reduction rather than novel judgment capability. Ensembling is Monte Carlo variance reduction: per-call variance drops as $1/k$ and the $k{=}1 \to k{=}8$ tie-rate collapse (20.4\% $\to$ 4.5\%) is its observable signature. Criteria injection does \emph{not} change per-response score variance (mean $\sigma_i{=}0.31$ vs 0.32; KS $p{=}0.88$) but sharpens \emph{between-response} discrimination (tie rate 4.5\% $\to$ 3.2\% at $k{=}8$; AUC for variance predicting incorrectness rises from 0.60 to 0.64). Per-response score variance itself provides a weak but measurable uncertainty signal (AUC${\approx}$0.60 for predicting incorrect judgments), which we examined as both a routing target and a diagnostic for judge error. Techniques that don't work (calibration, model routing, soft blending) inject new information or combine heterogeneous distributions; at high $k$, residual variance is small enough that these contribute noise rather than signal.

\paragraph{Conclusion.} On RewardBench~2, all four drop-in techniques we investigate improve over baseline. \textbf{Criteria injection} (near-free) and \textbf{ensembling} (with $k{=}3$ capturing ${\sim}$70\% of the gain) are the simplest and most cost-effective; together they reach up to \textbf{85.8\%} (+13.5pp over baseline), and we recommend them in practice for LLM-as-a-judge evaluation. These gains generalise across model providers, replicating on both OpenAI GPT and Anthropic Claude judges. Calibration context and adaptive model escalation also improve over baseline but at higher cost, and at high $k$ variance-reduction techniques become substitutes rather than complements, so stacking did not yield additive gains.

\FloatBarrier
\balance
\bibliographystyle{icml2026}
\bibliography{references}

\appendix
\section{Prompt Templates}
\label{sec:prompts}

\subsection{Base RB2 Prompt}
\label{sec:prompt-base}

\begin{lstlisting}[style=prompt]
### Task Description
Please act as an impartial judge and evaluate the quality of the response provided
by an AI assistant to the user query displayed below.

Notes:
1- Your evaluation should consider factors such as the helpfulness, relevance,
   accuracy, depth, creativity, and level of detail of the response.
2- Begin your evaluation by providing a short explanation.
3- Be as objective as possible. After providing your explanation, please rate the
   response on a scale of 1 to 10. For your rating, only give a number between 1
   and 10 (inclusive), do not use any markdown, and do not put any text after
   your final rating.

[Query]
{prompt}

[Response]
{completion}

[Your judgement]
\end{lstlisting}

\subsection{Criteria-Augmented Prompt}
\label{sec:prompt-criteria}

Identical to Appendix~\ref{sec:prompt-base} with the following appended to the generic-qualities list: \emph{``\ldots and level of detail of the response. \textbf{\{criterion\}}''}

\subsection{Calibration Context Prompt (Single Example)}
\label{sec:prompt-cal-single}

\begin{lstlisting}[style=prompt]
### Task Description
[Standard notes -- same as base prompt]

Here is a previously evaluated example from the same category for reference:

[Example Query]
{cal_prompt}

[Example Response]
{cal_response}

[Example Score: {cal_score}/10]

Now evaluate the following:

[Query]
{prompt}

[Response]
{completion}

[Your judgement]
\end{lstlisting}

\subsection{Calibration Context Prompt (Both Variant)}
\label{sec:prompt-cal-both}

Same structure as Appendix~\ref{sec:prompt-cal-single} but with two reference examples (one high-scoring, one low-scoring), demonstrating the scoring range to the judge.

\section{Per-Category Confidence Intervals}
\label{sec:per-cat-ci}

Table~\ref{tab:per-cat-ci} reports per-category accuracy with 95\% bootstrap confidence intervals (half-widths) for all conditions in Table~\ref{tab:results-main}. The smaller subsets (Math, $n{=}183$; Precise~IF, $n{=}159$) have notably wider CIs (typically $\pm$6--8pp), so per-category gains in these subsets should be interpreted with the corresponding uncertainty.

\begin{table*}[ht]
\centering
\caption{Per-category accuracy (\%) with 95\% bootstrap CI half-widths. Rows correspond to the conditions in Table~\ref{tab:results-main}.}
\label{tab:per-cat-ci}
\footnotesize
\begin{tabular}{lccccc}
\toprule
Condition & Factuality & Focus & Math & Precise IF & Safety \\
\midrule
\multicolumn{6}{l}{\textit{(a) Recommended techniques}} \\
\midrule
Baseline (full $k{=}1$)          & 76.4 ($\pm$3.8) & 70.1 ($\pm$4.1) & 61.2 ($\pm$6.8) & 34.0 ($\pm$7.6) & 87.3 ($\pm$3.2) \\
Criteria (full $k{=}1$)          & 77.9 ($\pm$3.7) & 72.3 ($\pm$4.0) & 73.2 ($\pm$6.4) & 32.1 ($\pm$7.3) & 90.6 ($\pm$2.8) \\
Ensemble (full $k{=}8$)          & 86.7 ($\pm$3.0) & 81.8 ($\pm$3.3) & 74.9 ($\pm$6.3) & 44.7 ($\pm$7.9) & 92.1 ($\pm$2.6) \\
\textbf{Criteria (full $k{=}8$)} & \textbf{89.1 ($\pm$2.8)} & \textbf{82.8 ($\pm$3.4)} & \textbf{79.2 ($\pm$5.9)} & 48.8 ($\pm$7.8) & \textbf{93.2 ($\pm$2.5)} \\
Mini model $k{=}8$               & 83.3 ($\pm$3.4) & 80.2 ($\pm$3.4) & 68.3 ($\pm$6.6) & 40.3 ($\pm$7.8) & 92.8 ($\pm$2.5) \\
Criteria (mini $k{=}8$)          & 85.5 ($\pm$3.1) & 82.2 ($\pm$3.4) & 72.7 ($\pm$6.4) & 41.2 ($\pm$7.7) & 94.9 ($\pm$2.1) \\
Nano model $k{=}8$               & 67.9 ($\pm$4.2) & 74.5 ($\pm$3.9) & 61.2 ($\pm$6.9) & 42.4 ($\pm$7.7) & 87.6 ($\pm$3.2) \\
Nano model $k{=}1$               & 45.6 ($\pm$4.6) & 51.0 ($\pm$4.4) & 45.1 ($\pm$7.2) & 26.3 ($\pm$6.8) & 74.9 ($\pm$4.2) \\
\midrule
\multicolumn{6}{l}{\textit{(b) Investigated techniques}} \\
\midrule
Calibration low ($k{=}1$)              & 78.9 ($\pm$3.6) & 71.5 ($\pm$3.9) & 65.6 ($\pm$7.0) & 32.5 ($\pm$7.1) & 89.9 ($\pm$2.9) \\
Calibration low ($k{=}8$)              & 86.7 ($\pm$3.0) & 80.6 ($\pm$3.4) & 75.4 ($\pm$6.2) & 46.9 ($\pm$7.7) & 93.0 ($\pm$2.4) \\
Calibration high ($k{=}1$)             & 77.3 ($\pm$3.8) & 68.4 ($\pm$4.0) & 67.8 ($\pm$6.7) & 34.4 ($\pm$7.3) & 87.7 ($\pm$3.1) \\
Calibration both ($k{=}1$)             & 77.3 ($\pm$3.8) & 71.1 ($\pm$4.0) & 65.6 ($\pm$6.9) & 31.9 ($\pm$7.0) & 88.8 ($\pm$3.0) \\
Calibration cross ($k{=}1$)            & 77.0 ($\pm$3.8) & 68.2 ($\pm$4.1) & 68.0 ($\pm$6.7) & 30.6 ($\pm$7.1) & 89.1 ($\pm$2.9) \\
Combined (full $k{=}8$)          & 87.6 ($\pm$2.9) & 80.6 ($\pm$3.5) & 77.6 ($\pm$6.0) & \textbf{52.5 ($\pm$7.9)} & 92.8 ($\pm$2.4) \\
\bottomrule
\end{tabular}
\end{table*}

Test-set rows from Table~\ref{tab:results-main} (Soft blend, Combined+blend, Variance-informed) are omitted here because the small held-out sample (${\sim}$340 examples, redistributed across categories) does not support reliable per-category CIs.

\section{Intersection-Based Accuracy}
\label{sec:intersection}

Conditions in Table~\ref{tab:results-main} have slightly different sample sizes ($N{=}1700$--$1746$) because Azure content filters refuse different prompts depending on prompt configuration. To verify that this refusal-driven sample variation does not bias the headline comparisons, we recompute accuracy on the \emph{intersection} of $N{=}1710$ examples that succeeded under all four main collections (baseline, criteria, calibration\_low, and combined). Table~\ref{tab:intersection} reports each headline condition's accuracy on its full sample alongside its accuracy on the intersection.

\begin{table*}[ht]
\centering
\caption{Headline conditions evaluated on their full sample versus the intersection of $N{=}1710$ examples shared across baseline, criteria, calibration\_low, and combined collections.}
\label{tab:intersection}
\footnotesize
\begin{tabular}{lrrrr}
\toprule
Condition & Full $N$ & Full Acc.\ & Intersection Acc.\ & $\Delta$ \\
\midrule
Baseline (full $k{=}1$)         & 1729 & 71.7\% & 71.7\% & $-$0.0 \\
Ensemble (full $k{=}8$)         & 1730 & 81.5\% & 81.6\% & $+$0.1 \\
\textbf{Criteria (full $k{=}8$)} & 1741 & \textbf{83.6\%} & \textbf{83.6\%} & $+$0.0 \\
Mini model $k{=}8$              & 1730 & 79.2\% & 79.1\% & $-$0.1 \\
Criteria (mini $k{=}8$)         & 1742 & 81.5\% & 81.3\% & $-$0.2 \\
Calibration low ($k{=}8$)             & 1744 & 81.7\% & 81.7\% & $+$0.0 \\
Combined (full $k{=}8$)         & 1746 & 82.6\% & 82.6\% & $-$0.0 \\
\bottomrule
\end{tabular}
\end{table*}

\section{Escalation Methods}
\label{sec:escalation-detail}

\paragraph{Motivation.} The mini-class model is ${\sim}3\times$ cheaper than the full-class model but somewhat less accurate. If we could identify in advance which examples the mini model will get wrong, we could route only those to the full model. We use the mini model's score variance as a proxy for example difficulty.

\paragraph{Data collection.} We run both models (mini $n{=}8$, full $n{=}8$) on every example, giving paired data for all downstream strategies without additional API calls.

\paragraph{Variance as a routing signal.} Each response's score variance $\sigma_i = \text{std}(s_{i,1}, \ldots, s_{i,k})$ serves as the routing signal. Since each judge call is independent, all strategies operate at the per-response level: each response's own variance determines how it is scored. The mini-vs-full correlation story is summarised in Section~\ref{sec:escalation}; for completeness, Figure~\ref{fig:variance-corr} below shows the cross-tier variance correlation. Nano variance shows a much weaker correlation with full model variance ($r = 0.106$) than mini ($r = 0.421$), suggesting that nano's scoring noise is largely independent and its uncertainty is not a useful proxy for the full model's.

\begin{figure}[t]
\centering
\includegraphics[width=\columnwidth]{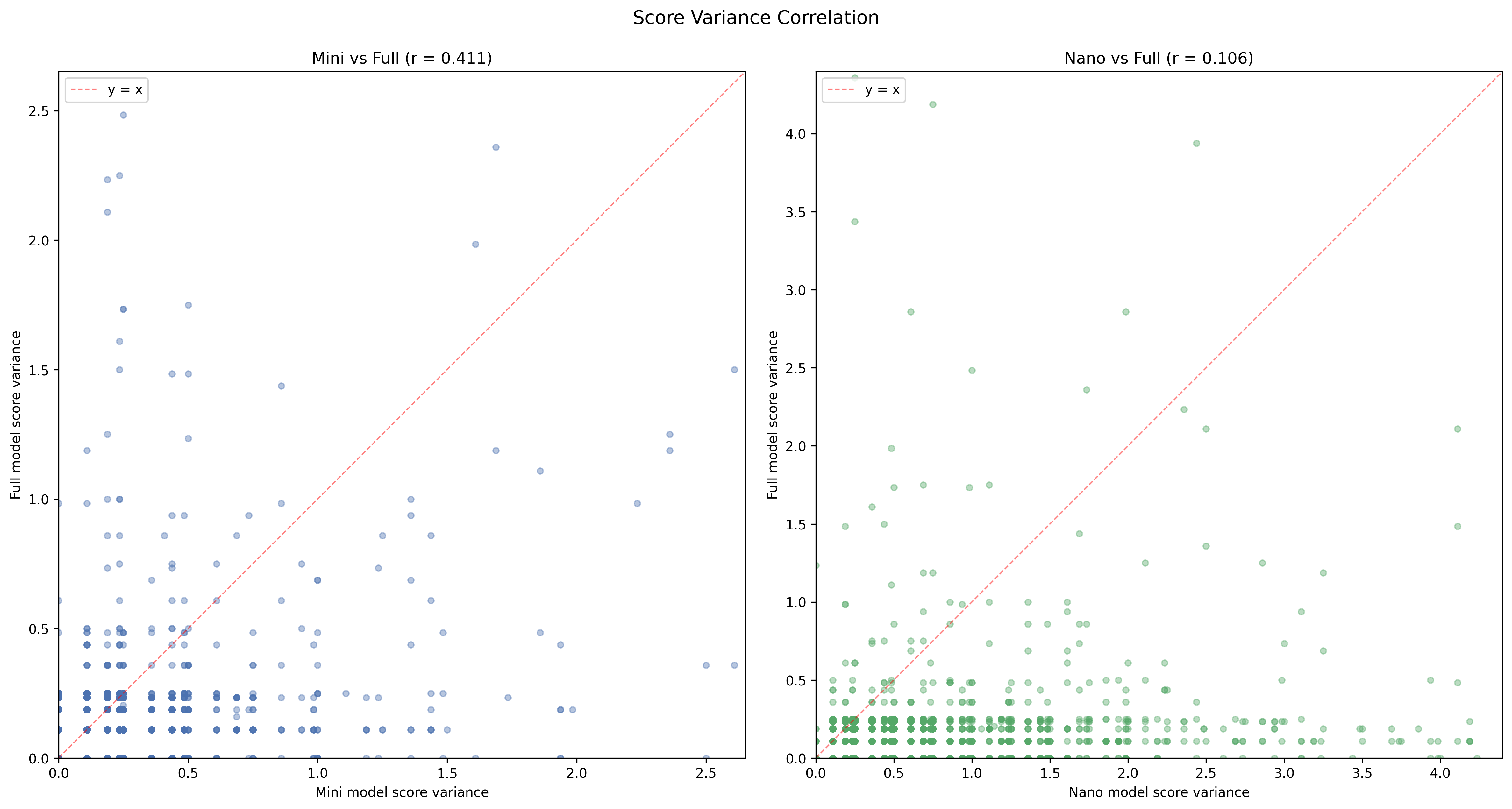}
\caption{Score variance correlation (chosen response). Left: mini vs full ($r = 0.421$). Right: nano vs full ($r = 0.106$).}
\label{fig:variance-corr}
\end{figure}

\subsection{Hard Variance Routing}
\label{sec:hard-routing}

If the mini model is uncertain about a response (high variance), discard its score and use the full model instead. For each individual response, use the full model score if the mini std exceeds a threshold $\theta$:

\begin{equation}
s_i^{\text{eff}} = \begin{cases}
s_i^{\text{full}} & \text{if } \text{std}(s_{i,1}^{\text{mini}}, \ldots, s_{i,k}^{\text{mini}}) \geq \theta \\
s_i^{\text{mini}} & \text{otherwise}
\end{cases}
\label{eq:hard-routing}
\end{equation}

The threshold $\theta$ is swept to trace the accuracy--cost tradeoff. Total cost scales with how often escalation is triggered: letting $p_{\text{esc}}$ denote the fraction of responses escalated,

\begin{equation}
C = C_{\text{mini}} + p_{\text{esc}} \cdot C_{\text{full}}
\label{eq:escalation-cost}
\end{equation}

where $C_{\text{mini}}$ and $C_{\text{full}}$ are the fixed costs of running all mini and full model calls respectively. Each value of $\theta$ yields one point in accuracy--cost space.

\begin{figure}[ht]
\centering
\includegraphics[width=\columnwidth]{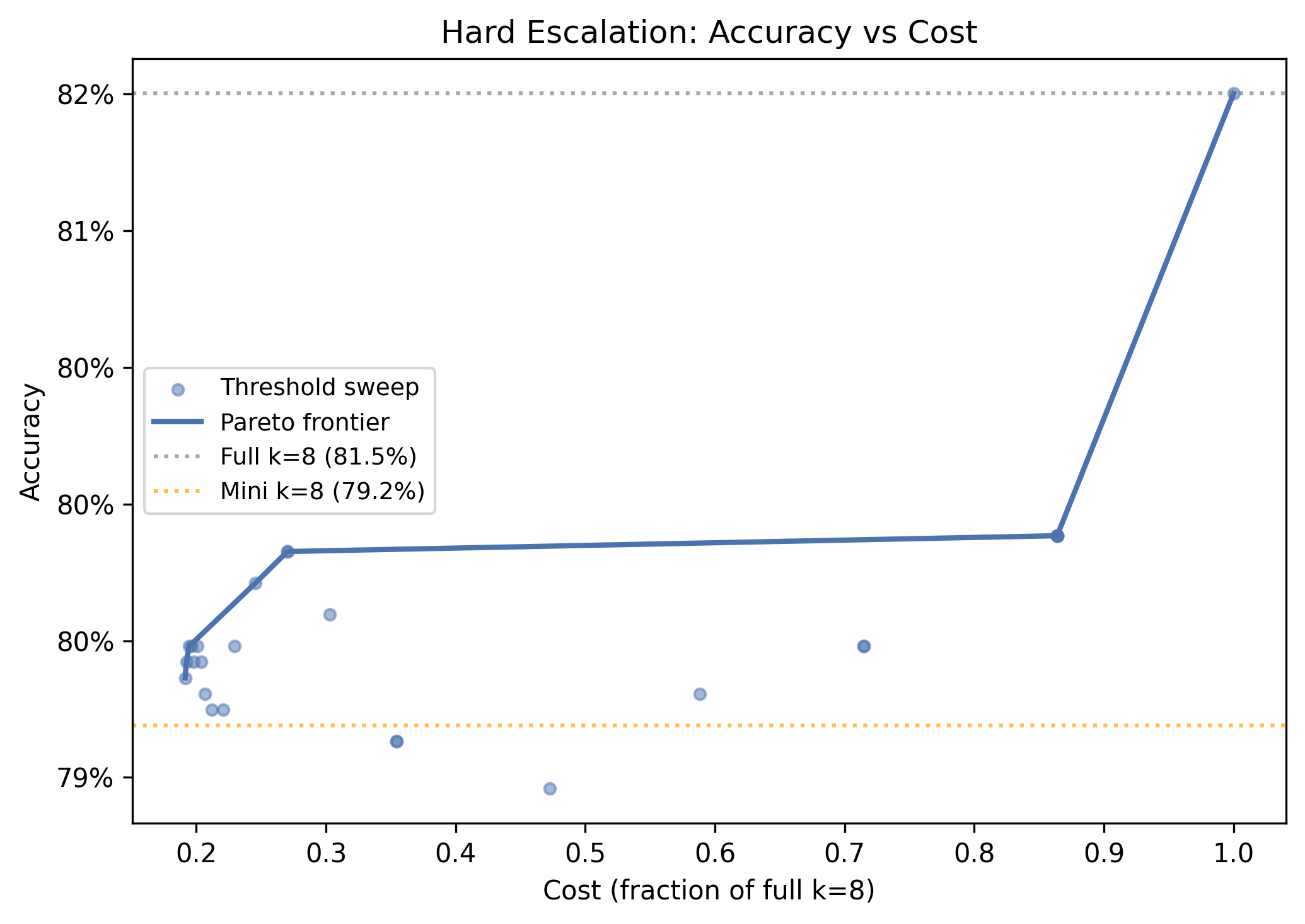}
\caption{Pareto frontier for per-response hard variance routing. Each point is a different threshold $\theta$. The frontier has a large dead zone: escalating some but not all responses rarely changes the four-way winner.}
\label{fig:escalation-pareto}
\end{figure}

\subsection{Soft Blending (Sigmoid)}
\label{sec:soft-blending}

If mini and full model scores are imperfectly correlated estimators of response quality, a weighted combination may have lower variance than either alone. We blend mini and full scores continuously using a per-response sigmoid weight:

\begin{align}
w_i(\sigma_i, m) &= \text{sigmoid}\left(10 \cdot (\sigma_i - m)\right) = \frac{1}{1 + e^{-10(\sigma_i - m)}} \label{eq:sigmoid-weight} \\
s_i^{\text{eff}} &= (1 - w_i) \cdot \bar{s}_i^{\text{mini}} + w_i \cdot \bar{s}_i^{\text{full}} \label{eq:soft-blend}
\end{align}

Each response's own variance $\sigma_i$ determines its blend weight independently. The midpoint $m$ controls where the transition from mini-dominant to full-dominant scoring occurs; steepness is fixed at 10. The optimal $m$ is found by sweeping over all unique per-response variance values.

\begin{figure}[ht]
\centering
\includegraphics[width=\columnwidth]{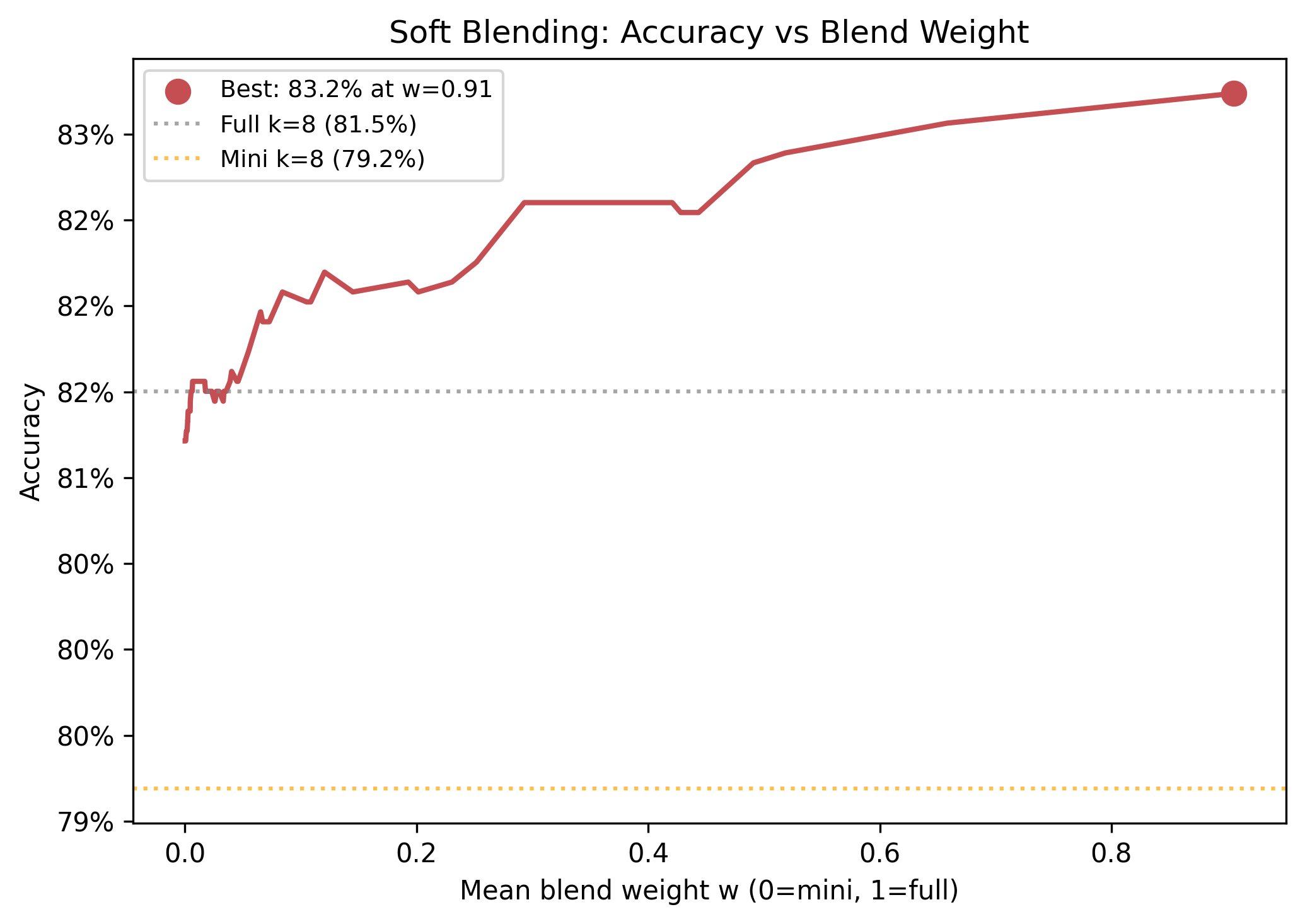}
\caption{Per-response soft blending accuracy vs mean blend weight $w$ (full dataset, in-sample). Accuracy peaks at 83.2\% near $w{=}0.91$ (mostly full model). On a held-out test set, the blend (80.2\%) does not beat full $k{=}8$ (81.5\%) --- the in-sample gain is partly midpoint overfitting.}
\label{fig:soft-blending-fig}
\end{figure}

\subsection{Variance-Informed Ensembling}
\label{sec:var-informed}

Hard routing and soft blending both use a fixed ensemble size ($k{=}8$) for both models. But Section~\ref{sec:ensemble} shows diminishing returns beyond $k{=}3$: most responses don't need 8 calls. We use each response's mini variance to determine $n_{\text{full},i}$, the number of full model calls for that response:

\begin{equation}
n_{\text{full},i}(\sigma_i) = \begin{cases}
1 & \text{if } \sigma_i \leq \sigma_1 \\
1 + \dfrac{(\sigma_i - \sigma_1)(n_{\max} - 1)}{\sigma_2 - \sigma_1} & \text{if } \sigma_1 < \sigma_i < \sigma_2 \\
n_{\max} & \text{if } \sigma_i \geq \sigma_2
\end{cases}
\label{eq:var-informed}
\end{equation}

Parameters $(\sigma_1, \sigma_2)$ are found by grid search over the 15th--95th percentile range of observed per-response variances.

\begin{figure}[ht]
\centering
\includegraphics[width=\columnwidth]{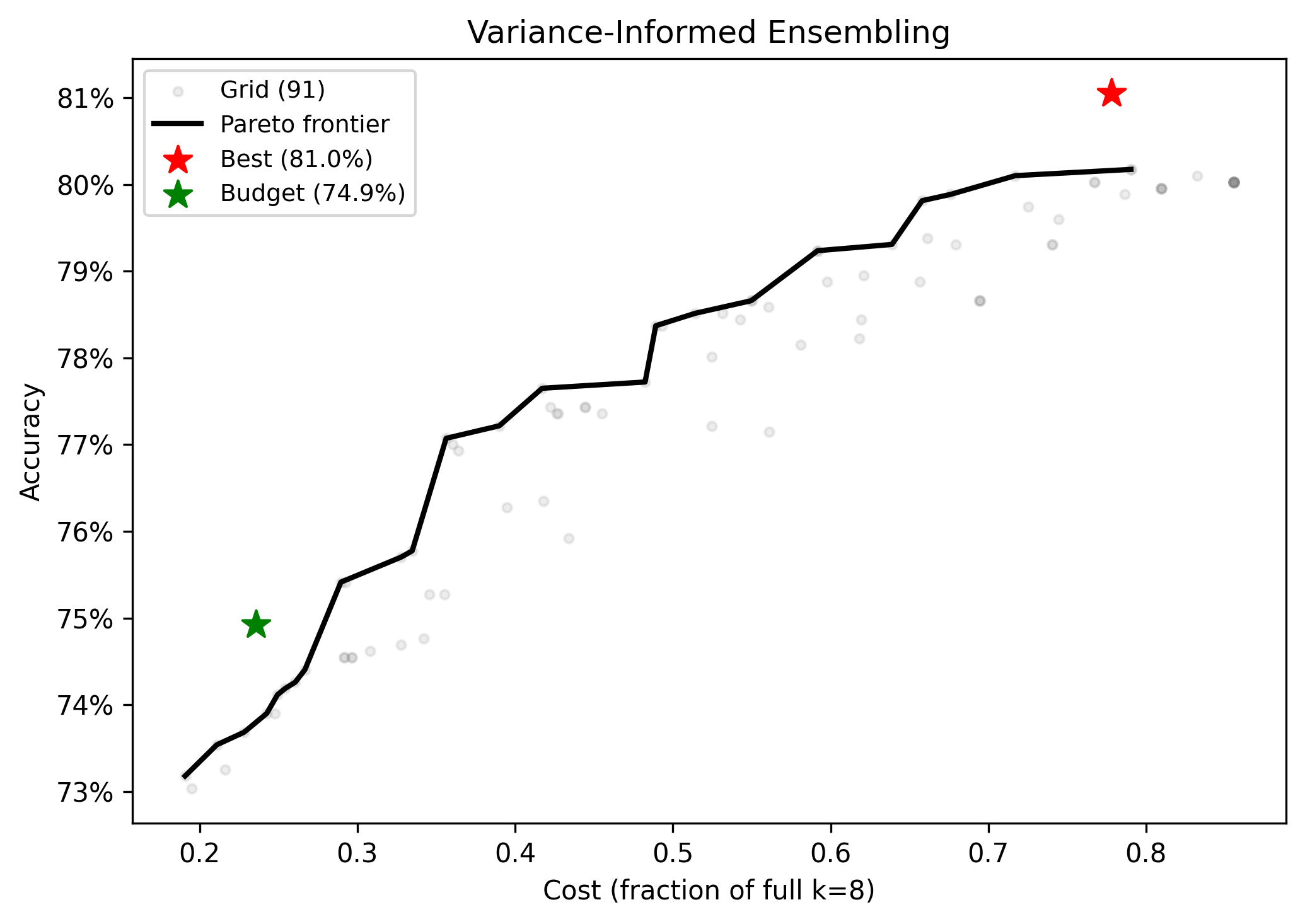}
\caption{Pareto frontier for per-response variance-informed ensembling. Gray points: full grid of $(\sigma_1, \sigma_2)$ configurations (train-set). Stars: test-set accuracy for budget-constrained ($\bar{n}_{\text{full}}{\leq}2$) and unconstrained configurations --- the budget-constrained point (74.9\%) is dominated by mini $k{=}8$ at comparable cost.}
\label{fig:var-informed}
\end{figure}

\subsection{Escalation Results}
\label{sec:escalation-results}

Table~\ref{tab:escalation-summary} compares escalation strategies against the $k{=}1$ full model baseline.

\begin{table}[ht]
\centering
\caption{Escalation strategy comparison. Test-set rows are the honest headline numbers; the in-sample soft-blend row is a tuned upper bound and should \emph{not} be compared on equal footing with the others.}
\label{tab:escalation-summary}
\small
\setlength{\tabcolsep}{4pt}
\begin{tabular}{lcc}
\toprule
Strategy & Acc.\ & vs $k{=}1$ \\
\midrule
$k{=}1$ full (baseline)        & 71.7\% & 1.0$\times$ \\
Full $k{=}8$                   & 81.5\% & 5.0$\times$ \\
Soft blend (in-sample)$^\dagger$ & 83.2\% & 6.1$\times$ \\
Soft blend (test)              & 80.2\% & 6.1$\times$ \\
Var-informed ($\bar{n}_{\text{full}}{\leq}2$, test) & 74.9\% & 1.6$\times$ \\
\bottomrule
\end{tabular}

\vspace{2pt}
{\footnotesize $^\dagger$ Tuned-parameter in-sample optimistic upper bound, not comparable with the other rows. The test-set row (80.2\%) is the honest generalisation number.}
\end{table}

Figure~\ref{fig:escalation-pareto} shows the hard variance routing Pareto frontier, with its characteristic dead zone in the middle --- escalating some but not all responses rarely changes the four-way winner, so meaningful operating points cluster at the mini-only or full-only extremes.

On the full dataset (in-sample), soft blending achieves 83.2\% vs 81.5\% for full $k{=}8$ (Figure~\ref{fig:soft-blending-fig}). However, on a held-out test set the base blend (80.2\%) does not beat full $k{=}8$ (81.5\%), indicating midpoint overfitting; we therefore do not report blend numbers as a positive result in the main paper.

The budget-constrained variance-informed variant restricts mean $n_{\text{full}} \leq 2.0$, achieving 74.9\% on the test set at 1.6$\times$ baseline cost (Figure~\ref{fig:var-informed}) --- barely above the baseline (71.7\%) and far below mini $k{=}8$ (79.2\% at 1.2$\times$).

\section{Calibration Detail}
\label{sec:calibration-detail}

Prompt listings for all calibration variants are in Appendix~\ref{sec:prompt-cal-single} and~\ref{sec:prompt-cal-both}. Table~\ref{tab:cal-results} reports calibration accuracy at $k{=}1$ and $k{=}8$; per-category CIs for the ``low'' variants are included in Appendix~\ref{sec:per-cat-ci}.

\begin{table}[ht]
\centering
\caption{Calibration accuracy at $k{=}1$ and $k{=}8$.}
\label{tab:cal-results}
\begin{tabular}{lcc}
\toprule
Variant & $k{=}1$ & $k{=}8$ \\
\midrule
Baseline (no calibration) & 71.7\% & 81.5\% \\
High           & 72.4\% & 81.0\% \\
Low            & 73.8\% & 81.7\% \\
Both           & 72.8\% & 81.5\% \\
Cross-category & 72.4\% & 81.7\% \\
\bottomrule
\end{tabular}
\end{table}

At $k{=}1$, all variants improve over baseline by +1--2pp. The ``low'' variant (showing a bad example) slightly outperforms ``high'' (73.8\% vs 72.4\%), possibly because the judge finds it easier to distinguish the target from a known-bad anchor. Cross-category calibration (72.4\%) performs identically to within-category, suggesting the benefit is anchoring rather than category-specific knowledge. At $k{=}8$, all variants fall within $\pm$0.2pp of ensembling (81.5\%): the $k{=}8$ ensemble already reduces scoring noise sufficiently that the anchoring benefit of calibration is redundant.

\section{Additional Analyses}
\label{sec:additional-analyses}

\subsection{Diminishing Returns Raw Numbers}

\begin{table}[ht]
\centering
\caption{Accuracy (\%) vs.\ ensemble size $k$ for three model tiers.}
\label{tab:diminishing-returns}
\begin{tabular}{cccc}
\toprule
$k$ & Full & Mini & Nano \\
\midrule
1 & 71.7 & 64.8 & 52.3 \\
2 & 77.3 & 73.0 & 60.8 \\
3 & 78.4 & 74.8 & 65.4 \\
4 & 79.4 & 76.8 & 68.1 \\
5 & 80.1 & 77.1 & 69.6 \\
6 & 80.9 & 77.7 & 70.0 \\
7 & 81.4 & 78.6 & 70.9 \\
8 & 81.5 & 79.2 & 71.4 \\
\bottomrule
\end{tabular}
\end{table}

\subsection{Temperature Sensitivity}
\label{sec:temperature}

\begin{figure}[t]
\centering
\includegraphics[width=\columnwidth]{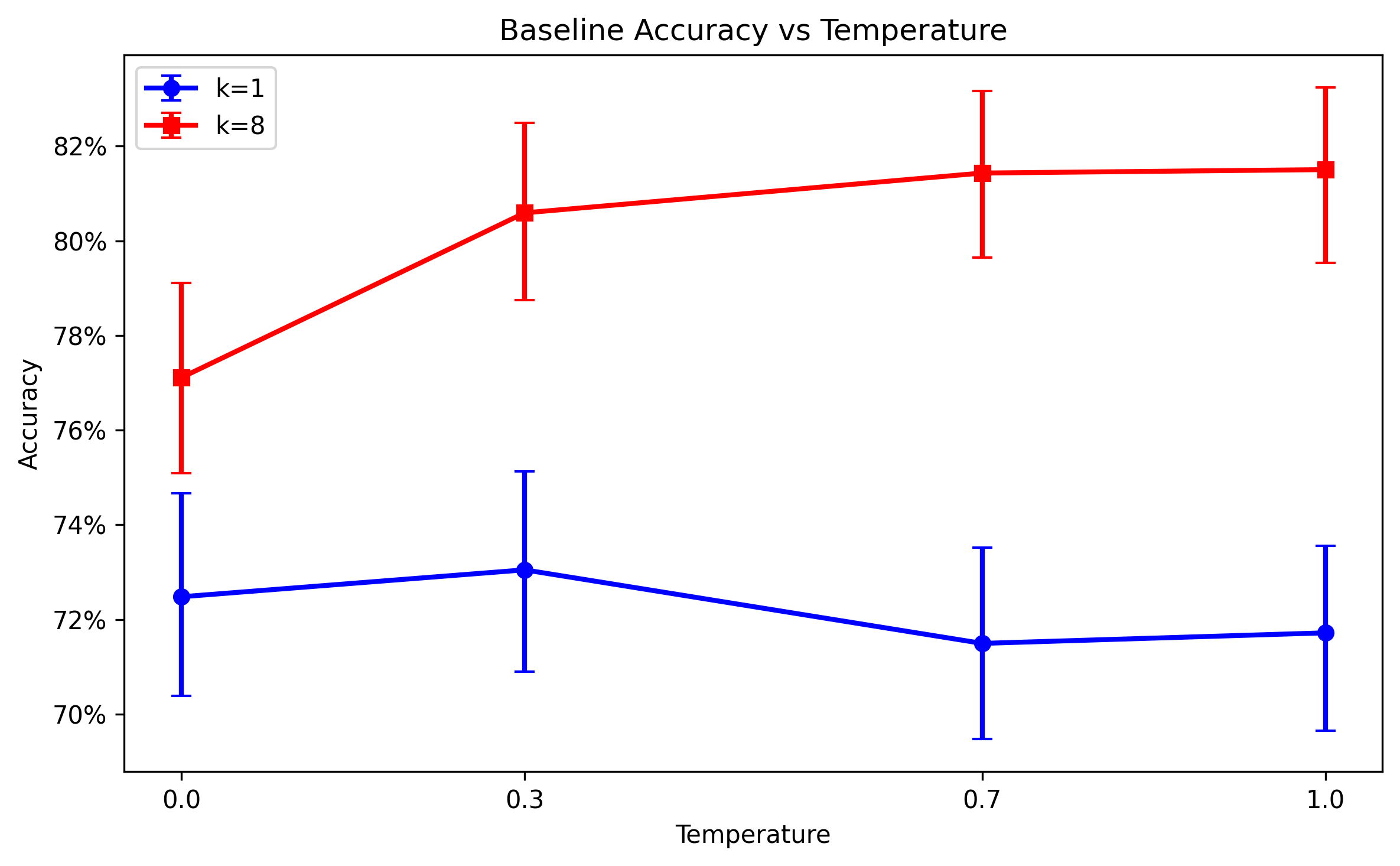}
\caption{Baseline accuracy vs temperature for $k{=}1$ and $k{=}8$ with 95\% bootstrap CIs.}
\label{fig:temperature}
\end{figure}

We sweep temperature across \{0.0, 0.3, 0.7, 1.0\} for the base prompt with the full model (Figure~\ref{fig:temperature}). At $k{=}1$, accuracy is stable across temperatures (71--73\%, CIs overlapping). At $k{=}8$, ensembling helps at all temperatures, but the gain increases with temperature: from +4.6pp at temperature~0 (72.5\% $\to$ 77.1\%) to +9.8pp at temperature~1.0 (71.7\% $\to$ 81.5\%). Even at temperature~0 there is a significant +4.6pp gap between $k{=}1$ (72.5\%) and $k{=}8$ (77.1\%), with non-overlapping confidence intervals --- \texttt{temperature=0} does not produce deterministic outputs in practice, likely due to floating-point non-determinism in GPU inference and the absence of a \texttt{seed} parameter. This is a useful finding for practitioners: even deployments that assume deterministic scoring at temperature~0 can benefit from ensembling.

\subsection{Mini--Full Convergence}
\label{sec:convergence}

The mini model ($n{=}8$) achieves 79.2\%, only 2.3pp below the full model ensemble (81.5\%). To understand the relationship, we measure how quickly mini's winner selection converges to the full model's as $k$ increases, using two statistics: \textbf{agreement} (fraction of examples where both models pick the same winner) and \textbf{Spearman rank correlation} $\rho$ between their mean-score vectors across the four responses.

\begin{table}[ht]
\centering
\caption{Mini--full model agreement and rank correlation vs.\ ensemble size.}
\label{tab:convergence}
\begin{tabular}{ccc}
\toprule
$k$ & Agreement & Rank corr ($\rho$) \\
\midrule
1 & 65.1\% & 0.761 \\
2 & 72.7\% & 0.771 \\
3 & 75.0\% & 0.780 \\
4 & 76.6\% & 0.779 \\
5 & 77.5\% & 0.781 \\
8 & 78.7\% & 0.785 \\
\bottomrule
\end{tabular}
\end{table}

Mini agreement reaches 78.7\% by $k{=}8$; we cannot distinguish from this data whether the ceiling reflects systematic disagreement or residual noise at a capability gap. Nano shows a qualitatively similar pattern but with a lower ceiling: agreement rises from ${\sim}$50\% at $k{=}1$ to ${\sim}$70\% at $k{=}8$, with rank correlation plateauing around 0.67 (vs 0.79 for mini). Visualisation is in Figure~\ref{fig:convergence} (main paper).

\end{document}